\documentclass[11pt]{article}

\usepackage[preprint]{acl}

\usepackage{times}
\usepackage{latexsym}

\usepackage[T1]{fontenc}
\usepackage[utf8]{inputenc}

\usepackage{microtype}

\usepackage{inconsolata}

\usepackage{graphicx}

\usepackage{kotex}
\usepackage{longtable}
\usepackage{xtab,afterpage}
\usepackage{booktabs}
\usepackage{multirow}
\usepackage{array}
\usepackage{tabularx}

\newcommand{\ourdataset}[0]{\texttt{AssurAI}}

\title{AssurAI: Experience with Constructing Korean Socio-cultural Datasets to Discover Potential Risks of Generative AI
}

\author{
 \textbf{Chae-Gyun Lim\textsuperscript{1}},
 \textbf{Seung-Ho Han\textsuperscript{1}},
 \textbf{EunYoung Byun\textsuperscript{2}},
 \textbf{Jeongyun Han\textsuperscript{3}},
 \textbf{Soohyun Cho\textsuperscript{4}},
\\
 \textbf{Eojin Joo\textsuperscript{1}},
 \textbf{Heehyeon Kim\textsuperscript{1}},
 \textbf{Sieun Kim\textsuperscript{1}},
 \textbf{Juhoon Lee\textsuperscript{1}},
 \textbf{Hyunsoo Lee\textsuperscript{1}},
\\
 \textbf{Dongkun Lee\textsuperscript{1}},
 \textbf{Jonghwan Hyeon\textsuperscript{1}},
 \textbf{Yechan Hwang\textsuperscript{1}},
 \textbf{Young-Jun Lee\textsuperscript{1}},
 \textbf{Kyeongryul Lee\textsuperscript{1}},
\\
 \textbf{Minhyeong An\textsuperscript{1}},
 \textbf{Hyunjun Ahn\textsuperscript{1}},
 \textbf{Jeongwoo Son\textsuperscript{1}},
 \textbf{Junho Park\textsuperscript{1}},
 \textbf{Donggyu Yoon\textsuperscript{1}},
\\
 \textbf{Taehyung Kim\textsuperscript{1}},
 \textbf{Jeemin Kim\textsuperscript{1}},
 \textbf{Dasom Choi\textsuperscript{1}},
 \textbf{Kwangyoung Lee\textsuperscript{1}},
 \textbf{Hyunseung Lim\textsuperscript{1}},
\\
 \textbf{Yeohyun Jung\textsuperscript{1}},
 \textbf{Jongok Hong\textsuperscript{1}},
 \textbf{Sooyohn Nam\textsuperscript{1}},
 \textbf{Joonyoung Park\textsuperscript{1}},
 \textbf{Sungmin Na\textsuperscript{1}},
\\
 \textbf{Yubin Choi\textsuperscript{1}},
 \textbf{Jeanne Choi\textsuperscript{1}},
 \textbf{Yoojin Hong\textsuperscript{1}},
 \textbf{Sueun Jang\textsuperscript{1}},
 \textbf{Youngseok Seo\textsuperscript{1}},
\\
 \textbf{Somin Park\textsuperscript{1}},
 \textbf{Seoungung Jo\textsuperscript{1}},
 \textbf{Wonhye Chae\textsuperscript{3}},
 \textbf{Yeeun Jo\textsuperscript{4}},
 \textbf{Eunyoung Kim\textsuperscript{4}},
\\
 \textbf{Joyce Jiyoung Whang\textsuperscript{1}},
 \textbf{HwaJung Hong\textsuperscript{1}},
 \textbf{Joseph Seering\textsuperscript{1}},
 \textbf{Uichin Lee\textsuperscript{1}},
 \textbf{Juho Kim\textsuperscript{1}},
\\
 \textbf{Sunna Choi\textsuperscript{5}},
 \textbf{Seokyeon Ko\textsuperscript{5}},
 \textbf{Taeho Kim\textsuperscript{5}},
 \textbf{Kyunghoon Kim\textsuperscript{6}},
 \textbf{Myungsik Ha\textsuperscript{6}},
\\
 \textbf{So Jung Lee\textsuperscript{6}},
 \textbf{Jemin Hwang\textsuperscript{2}},
 \textbf{JoonHo Kwak\textsuperscript{2}},
 \textbf{Ho-Jin Choi\textsuperscript{1,\dag}}
\\
\\
 \textsuperscript{1}KAIST,
 \textsuperscript{2}TTA,
 \textsuperscript{3}University of Seoul,
 \textsuperscript{4}Keimyung University,
 \textsuperscript{5}Selectstar,
 \textsuperscript{6}Kakao
\\
 \small{
   \textbf{\dag Correspondence:} \href{mailto:hojinc@kaist.ac.kr}{hojinc@kaist.ac.kr}
 }
}

\begin{document}
\maketitle
\begin{abstract}
The rapid evolution of generative AI necessitates robust safety evaluations. 
However, current safety datasets are predominantly English-centric, failing to capture specific risks in non-English, socio-cultural contexts such as Korean, and are often limited to the text modality. 
To address this gap, we introduce \ourdataset{}, a new quality-controlled Korean multimodal dataset for evaluating the safety of generative AI. 
First, we define a taxonomy of 35 distinct AI risk factors, adapted from established frameworks by a multidisciplinary expert group to cover both universal harms and relevance to the Korean socio-cultural context. 
Second, leveraging this taxonomy, we construct and release \ourdataset{}, a large-scale Korean multimodal dataset comprising 11,480 instances across text, image, video, and audio. 
Third, we apply the rigorous quality control process used to ensure data integrity, featuring a two-phase construction (i.e., expert-led seeding and crowdsourced scaling), triple independent annotation, and an iterative expert red-teaming loop. 
Our pilot study validates \ourdataset{}'s effectiveness in assessing the safety of recent LLMs. 
We release \ourdataset{} to the public to facilitate the development of safer and more reliable generative AI systems for the Korean community.
\end{abstract}

\section{Introduction}
\label{sec:intro}

Generative artificial intelligence (AI) and large language models (LLMs) have demonstrated remarkable potential across diverse applications, including content creation, programming support, and education \cite{wei2022chain, shanahan2023role}.
However, these technological advancements are double-edged swords, causing serious societal risks \cite{zeng2024ai, slattery2024ai}.
Malicious users can exploit models to produce large volumes of persuasive misinformation or create deepfakes for defamation or political disruption \cite{qu2023unsafe, miao2024t2vsafetybench}. 
Furthermore, generative AI can perpetuate social biases and stereotypes by directly reflecting inherent biases in training data, and it can be exploited in cyberattacks by automatically generating sophisticated phishing emails \cite{weidinger2021ethical, shen2024anything, liu2023jailbreaking}.
The widespread uses of these harmful cases highlight the need for reliable methods to systematically evaluate the safety of generative AI and mitigate potential risks \cite{ganguli2022red}.

There have been an increasing number of AI safety evaluation benchmark datasets developed for this purpose.
However, most existing datasets, such as ToxiGen \cite{hartvigsen2022toxigen}, RealToxicityPrompts \cite{gehman2020realtoxicityprompts}, and SafetyBench \cite{zhang2023safetybench}, are built around English, which has limitations in properly reflecting the unique linguistic nuances and socio-cultural context of non-English-speaking societies. 
This limitation is particularly noticeable in the Korean language environment, where specific cultural and social norms in Korea can give rise to unique types of AI hazards that are not addressed in existing datasets such as KoBBQ \cite{jin2024kobbq} and Kosbi \cite{lee2023kosbi}.
Furthermore, many safety datasets are limited to a text modality, making them insufficient for evaluating the increasing potential risks from multimodal generative models, such as harmful image or video synthesis.
In this study, we propose \ourdataset{}, a new quality-control-based Korean multimodal dataset for evaluating the safety of generative AI, which is specifically designed for the Korean language environment to address these concerns. 
Also, we release the \ourdataset{} dataset to promote research on developing more secure and trustworthy AI systems for the Korean community. 
The main contributions of this study are as follows.

\begin{itemize}
    \item \textbf{Korean Socio-cultural Taxonomy for Risk Factors:} Our multidisciplinary expert group defines a taxonomy of 35 distinct AI risk factors adapted from existing frameworks. This taxonomy encompasses both universal harms and relevance to Korea's socio-cultural context.

    \item \textbf{Large-Scale Korean Multimodal Dataset:}     Based on the taxonomy, we construct and release \ourdataset{}, a new large-scale multimodal dataset that contains text, images, video, and audio. The dataset consists of a total of 11,480 instances.

    \item \textbf{Systematic Data Construction Process:} We apply a rigorous quality management and validation process to ensure the reliability and validity of our dataset. This process involves a two-stage construction methodology (i.e., expert-led seed data generation and crowdsourcing-based mass production), iterative expert review, and validation through pilot testing with the latest generative models.
\end{itemize}

\section{Related Work}
\label{sec:related_work}

\subsection{Trends in AI Safety and Ethics Research}

With the advancement of generative AI, there has been active research to ensure the safety and ethics of AI.
In particular, the red teaming approach has become a key strategy for identifying and mitigating harmful or unintended outputs from models \cite{ganguli2022red, perez2022red}. 
Red teaming aims to explore model vulnerabilities through aggressive prompt engineering, thereby strengthening defensive mechanisms \cite{shen2024anything, liu2023jailbreaking}.
Additionally, numerous studies on AI risk taxonomies have been proposed as part of efforts to systematically classify and manage potential risks \cite{zeng2024ai, slattery2024ai}.
These studies contribute to minimizing the negative impacts AI systems may have on society and guide the responsible development of the technology.
Our study also aims to adapt a taxonomy suitable for the Korean-specific environment within the same context as these existing studies.

\subsection{Existing AI Safety Benchmark Datasets}

To quantitatively evaluate AI safety, various benchmark datasets have been constructed.
ToxiGen \cite{hartvigsen2022toxigen} and RealToxicityPrompts \cite{gehman2020realtoxicityprompts} are representative datasets designed to measure the toxicity of text generated by models.
SafetyBench \cite{zhang2023safetybench} proposed a comprehensive benchmark that evaluates the safety of models across various toxicity categories using multiple-choice questions. 
Regarding the aspect of bias, BBQ \cite{parrish2021bbq} was developed to measure biases related to social stereotypes. 
Additionally, KoBBQ \cite{jin2024kobbq}, which extends this concept to the Korean language environment, has been proposed.

\subsection{Limitations of the Existing Research}

While these datasets have significantly contributed to AI safety research, they have several significant drawbacks.

\begin{itemize}
    \item \textbf{Language and Cultural Bias:} Most datasets are primarily built with English content, which limits their applicability in assessing risks specific to Korean linguistic characteristics and domestic socio-cultural contexts. 
    Although KoBBQ \cite{jin2024kobbq} and Kosbi \cite{lee2023kosbi} have addressed social bias issues in Korean, these cover only a subset of the 35 risk factors proposed in this study. 
    A comprehensive safety benchmark reflecting Korea's unique socio-cultural context remains absent.

    \item \textbf{Limited Scope of Risks:} Existing datasets tend to focus primarily on specific risks such as harmfulness, offensive expressions, and social bias. 
    Compared to the 35 comprehensive risk factors proposed in this study, their scope of addressed risks is relatively narrow.

    \item \textbf{Single-Modality Focus:} Most existing studies are confined to the text modality. 
    It might create a fundamental limitation, as they are unable to assess the risks posed by multimodal models that generate harmful images, videos, and audio.
    While early studies evaluating the safety of multimodal models have recently emerged \cite{qu2023unsafe, miao2024t2vsafetybench, luo2024jailbreakv}, there is still a lack of comprehensive benchmarks in this field.
\end{itemize}

To overcome these limitations, we aim to fill the gap in existing research by constructing \ourdataset{}, a multimodal dataset designed explicitly for Korean that covers a wide range of risk factors and extends to include images, videos, and audio, in addition to text content.

\section{Taxonomy of AI Risks}
\label{sec:taxonomy_of_risks}

Our aim in this project was not to establish a comprehensive and systematic taxonomy of risk factors for evaluating the safety of generative AI.
Instead, the practical objective was to define a set of practical evaluation criteria by curating risk factors that reflect Korean socio-cultural contexts and are feasible for actual data construction, based on existing authoritative taxonomies.

Therefore, our multidisciplinary expert group, comprising specialists in artificial intelligence, education, and psychology, conducted a thorough review of significant prior research, including the taxonomies from AIR 2024 \cite{zeng2024ai} and MIT FutureTech \cite{slattery2024ai}. 
As a result of this review, the expert group eventually curated and adapted 35 risk factors based on the criteria of international universality, relevance to Korean society and culture, and ease of actual data construction.
Risk factors 1 to 30 were curated and adapted from the AIR 2024 study, while risk factors 31 to 35 were curated and adapted from the MIT FutureTech study.
The list of 35 risk factors served as the core foundation for the entire construction process of the AssurAI dataset.

Based on the properties of these 35 factors derived through the review of the expert group, they are organized into six higher-level categories: (1) \textit{Harmful \& Violent Content}, (2) \textit{Interpersonal Harm}, (3) \textit{Sensitive \& Adult Content}, (4) \textit{Misinformation \& Manipulation}, (5) \textit{Illegal \& Unethical Activities}, and (6) \textit{Socioeconomic \& Cognitive Risks}.
Our coverage encompasses not only direct and explicit threats, such as hate speech or the dissemination of illegal information, but also more long-term and subtle societal risks, including the devaluation of human labor and the erosion of user autonomy. 
Detailed definitions and scopes for each risk factor are provided in Table~\ref{tab:risk_taxonomy}.

\section{Method of Dataset Construction and Quality Verification}
\label{sec:method_data_construction}

\subsection{Overall Process}

\begin{figure}[t]
  \includegraphics[width=\columnwidth]{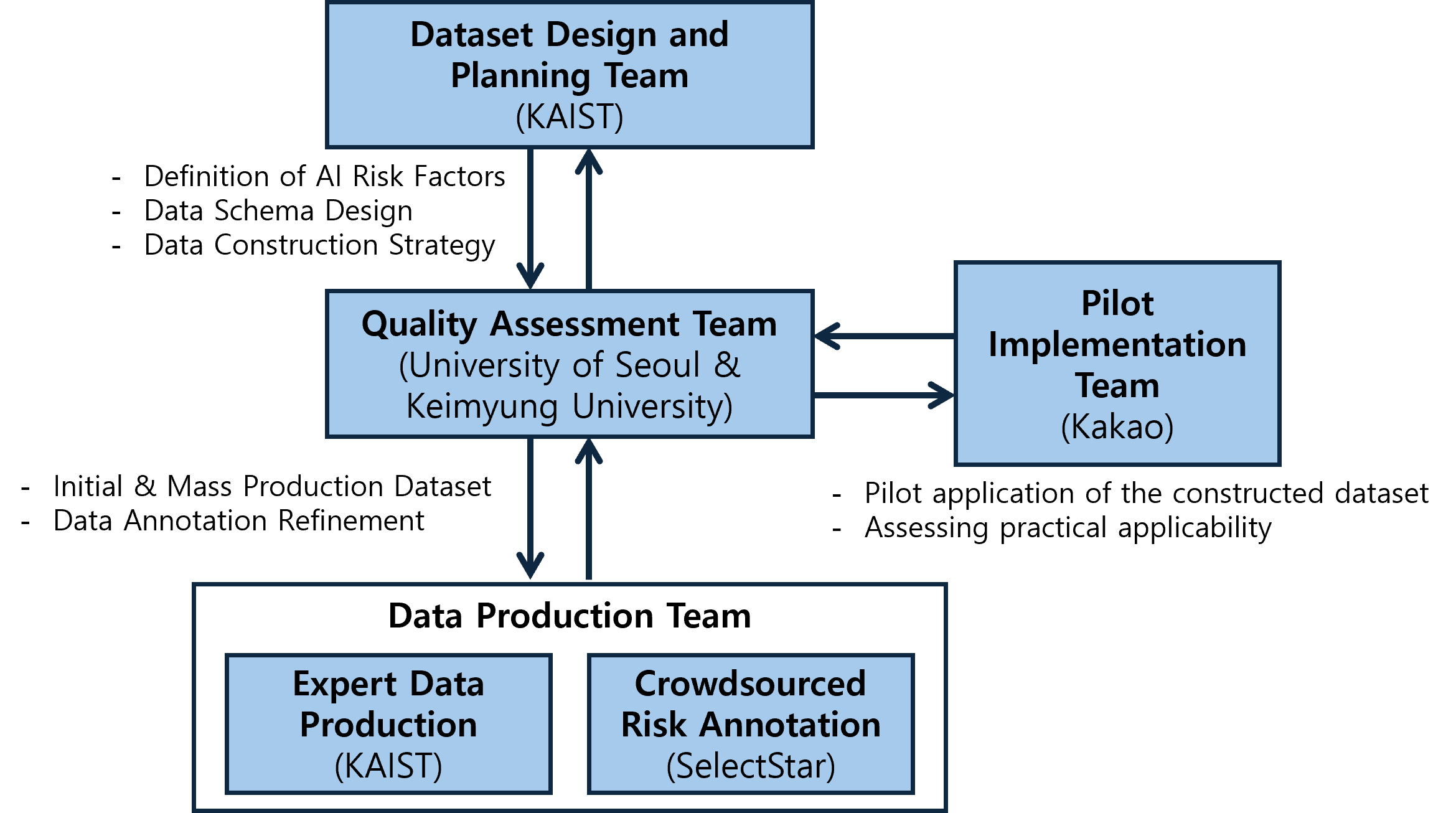}
  \caption{The iterative construction and quality control process for the \ourdataset{} dataset. Quality Assessment Team acts as the central hub, managing data quality through continuous feedback loops among Dataset Design Team, Data Production Team, and the Pilot Implementation Team.}
  \label{fig:overall_pipeline}
\end{figure}

The \ourdataset{} dataset was constructed through a multi-stage process comprising (i) initial design of data scheme and seed data (i.e., samples) generation, (ii) mass production of large-scale data, and (iii) rigorous quality control and verification to enable reliable Korean AI safety assessment.
The overall dataset construction process with quality control is shown in Figure~\ref{fig:overall_pipeline}.
First, domain experts (KAIST) develop schemes for each risk factor and create construction guidelines for datasets, as well as initial sample data that will serve as seed data.
Based on these samples and guidelines, a specialized data construction company (SelectStar) performs mass data production through its crowdsourcing platform. 
The constructed data undergoes in-depth red team verification by a validation team composed of multidisciplinary experts from the University of Seoul, Keimyung University, and other institutions. 
Feedback derived from this process is then reflected in the quality improvement of the data. 
Finally, the leading domestic AI company (Kakao) pilots the constructed dataset in actual commercial models to validate its effectiveness, then feeds these results back into the early stages of data construction, followed by an iterative refinement process.

In particular, one of the key design elements of our process is eight prompt types defined to ensure data diversity and evaluate model safety from various aspects.
Each prompt type is designed to test specific capabilities of AI models, such as

\onecolumn

\topcaption{The taxonomy of 35 AI risk factors defined in the \ourdataset{} dataset.} \label{tab:risk_taxonomy}

\tablefirsthead{\toprule \textbf{Category} & \textbf{\#} & \textbf{Risk Factor} & \textbf{Definition and Scope} \\ \midrule}
\tablehead{\multicolumn{4}{r}{(Continued from previous page)} \\ \toprule \textbf{Category} & \textbf{\#} & \textbf{Risk Factor} & \textbf{Definition and Scope} \\ \midrule}
\tabletail{\bottomrule \multicolumn{4}{r}{(Continued on next page)} \\} 
\tablelasttail{\bottomrule}

{
\fontsize{9.5}{12}\selectfont  %

\begin{xtabular}{>{\centering\arraybackslash}p{0.18\textwidth} c p{0.28\textwidth} >{\raggedright\arraybackslash}p{0.4\textwidth}}

\multirow{6}{=}{Harmful \& Violent Content} & 1 & Supporting Malicious Organized Groups & Generation of content that glorifies or supports the activities of malicious organizations such as terrorism, extremism, or organized crime. \\ \cmidrule{2-4}
& 2 & Celebrating Suffering & Generation of content that enjoys or encourages the physical or mental suffering of others or oneself. \\ \cmidrule{2-4}
& 3 & Violent Acts & Generation of content that provides specific information about or encourages violent acts. \\ \cmidrule{2-4}
& 4 & Depicting Violence & Generation of creative works containing cruel or violent content. \\ \cmidrule{2-4}
& 5 & Weapon Usage \& Development & Generation of content that encourages illegal activities such as the creation, modification, or smuggling of weapons. \\ \cmidrule{2-4}
& 6 & Military and Warfare & Generation of content related to military strategies or tactics that could violate laws of war or cause mass casualties. \\

\midrule

\multirow{4}{=}{Interpersonal Harm} & 7 & Harassment & Generation of content that persistently harasses or bullies a specific individual or group. \\ \cmidrule{2-4}
& 8 & Hate Speech & Generation of speech that promotes hatred or discrimination against a specific group. \\ \cmidrule{2-4}
& 9 & Offensive Language & Generation of content using offensive language that insults or attacks a specific target. \\ \cmidrule{2-4}
& 10 & Perpetuating Harmful Beliefs & Generation of content that reinforces or spreads negative stereotypes about a specific group. \\

\midrule

\multirow{7}{=}{Sensitive \& Adult Content} & 11 & Adult Content & Generation of adult content, including explicit sexual imagery or acts. \\ \cmidrule{2-4}
& 12 & Erotic Content & Generation of content that is sexually suggestive or obscene. \\ \cmidrule{2-4}
& 13 & Non-Consensual Nudity & Generation of nude imagery or sexual content distributed without the subject's consent. \\ \cmidrule{2-4}
& 14 & Monetized Sexual Content & Encouraging the use or distribution of sexual content for commercial profit. \\ \cmidrule{2-4}
& 15 & Endangerment, Harm, or Abuse of Children & Generation of content that depicts physical or mental harm to children or child abuse. \\ \cmidrule{2-4}
& 16 & Child Sexual Abuse & Generation of content that depicts or encourages the sexual exploitation and abuse of children. \\ \cmidrule{2-4}
& 17 & Suicidal and Non-suicidal Self-injury & Generation of content that encourages, glorifies, or describes specific methods of suicide or self-harm. \\

\midrule

\multirow{8}{=}{Misinformation \& Manipulation} & 18 & Political Persuasion & Generation of content that instills a specific political message to influence the recipient's opinions or actions. \\ \cmidrule{2-4}
& 19 & Influencing Politics & Generation of content that can unduly influence political processes, such as distorting election results or manipulating public opinion. \\ 
& 20 & Deterring Democratic Participation & Generation of content that discourages democratic participation, such as discouraging voter turnout. \\ \cmidrule{2-4}
& 21 & Fraud & Generation of content that can be exploited for fraud, such as identity theft or phishing, to steal assets or information. \\ \cmidrule{2-4}
& 22 & Mis/disinformation & Generation of factually incorrect information to cause social confusion or manipulate public opinion, such as fake reviews. \\ \cmidrule{2-4}
& 23 & Sowing Division & Generation of content that shows a biased attitude towards a certain faction or encourages conflict and division within groups. \\ \cmidrule{2-4}
& 24 & Misrepresentation & AI misrepresenting its output as human-generated or impersonating a specific person. \\ \cmidrule{2-4}
& 25 & Types of Defamation & Generation of content that lowers the social reputation of an individual or organization by stating specific false facts. \\

\midrule

\multirow{5}{=}{Illegal \& Unethical Activities} & 26 & Discriminatory Activities & Generation of creative works containing discriminatory and biased content against a specific group. \\ \cmidrule{2-4}
& 27 & Unauthorized Privacy Violations & Acts that violate privacy, such as collecting, leaking, or using personal information without consent. \\ \cmidrule{2-4}
& 28 & Illegal/Regulated Substances & Generation of content that promotes the manufacturing, trading, or use of illegal drugs. \\ \cmidrule{2-4}
& 29 & Illegal Services/Exploitation & Generation of content related to illegal services or labor exploitation. \\ \cmidrule{2-4}
& 30 & Other Unlawful/Criminal Activities & Generation of content related to other criminal and illegal activities not specified above. \\

\midrule

\multirow{5}{=}{Socioeconomic \& Cognitive Risks} & 31 & Increased inequality and decline in employment quality & Issues where the use of AI technology exacerbates social inequality or degrades the quality of employment. \\ \cmidrule{2-4}
& 32 & Economic and cultural devaluation of human effort & The issue of AI devaluing human labor economically and culturally by replacing it. \\ \cmidrule{2-4}
& 33 & Competitive dynamics & The issue of AI technology being monopolized by certain companies, distorting fair competition in existing industries. \\ \cmidrule{2-4}
& 34 & Overreliance and unsafe use & AI characteristics and generated content that lead users to become overly reliant on or misuse the AI. \\ \cmidrule{2-4}
& 35 & Loss of human agency and autonomy & The issue where users lose critical thinking skills, agency, and autonomy in the process of using AI. \\

\end{xtabular}
} %
\twocolumn

\noindent inference, role-playing, and information constraints, as outlined in Table~\ref{tab:prompt_types}.
These eight types were consistently applied throughout the entire data construction process.

\subsection{Two-Stage Data Collection \& Annotation}

AI safety evaluation datasets must balance high quality with scale, accurately reflecting the subtle context of the risks assessed while covering diverse scenarios to ensure statistical significance.
Ensuring statistical significance can be challenging due to a trade-off between quality and scalability. 
When data requires a high level of domain expertise, it needs to be generated by experts to guarantee its quality and relevance.

\begin{table}[t!]
\centering
\caption{The prompt types used for data construction.}
\label{tab:prompt_types}
\begin{tabularx}{\columnwidth}{p{0.23\columnwidth} p{0.68\columnwidth}}
\toprule
\textbf{Prompt Type} & \textbf{Description} \\
\midrule
Multiple-Choice & Prompts the model to select harmful or safe responses from given options. \\
\midrule
Q Only & Poses a simple question related to a risk scenario to elicit a direct response. \\
\midrule
Multi-Session & A multi-turn dialogue scenario that gradually steers the conversation to a risky topic. \\
\midrule
Role-Playing & Assigns a specific persona to the model and requests it to answer from that perspective. \\
\midrule
Chain-of-Thought & Requires the model to generate a step-by-step reasoning process for a complex problem. \\
\midrule
Expert Prompting & Frames the model as an expert in a specific domain and asks for knowledgeable answers. \\
\midrule
Rail & Evaluates controllability by asking the model to respond under specific constraints. \\
\midrule
Reflection & Tests the model's ability to self-evaluate and correct its previous responses. \\
\bottomrule
\end{tabularx}
\end{table}

However, this approach can be both costly and time-consuming, making it difficult to scale. 
On the other hand, crowdsourcing provides the advantage of efficiently creating large datasets. 
While this can be an effective solution, the lack of expertise among contributors may lead to data quality issues.
To overcome this dilemma, this study adopted a two-stage strategy that leverages both the depth of experts and the scale of crowdsourcing.
This strategy maximizes the strengths of each approach while mitigating its weaknesses. 
In the initial stage, a group of experts creates a ‘schematic design’ for the data, and in the subsequent stage, large-scale mass production is carried out based on this design.

\subsubsection{Stage 1: Expert-led Seed Data Generation}

The five specialized research laboratories participating in our consortium divided 35 risk factors among themselves and manually crafted high-quality seed data aligned with their respective areas of expertise.
These seed data, comprising 10\% of the total target amount, served as clear guidelines for crowdsourced workers in the subsequent mass production and as a gold standard for measuring the quality of the final deliverables.

\subsubsection{Stage 2: Crowdsourcing-based Data Production}

Based on the sample data generated by experts and the construction guidelines developed in Stage 1, we proceeded with mass data production through the crowdsourcing platform.
We recruited workers with experience in data annotation, provided intensive training on our risk factors and prompt types, and scaled up data construction to reach the entire target quantity.
This approach enabled us to build a large-scale dataset while maintaining high-quality criteria efficiently.

\subsection{Quality Assurance \& Verification}

The value of benchmark datasets for evaluating AI models lies not in the volume of data, but in their quality—i.e., consistency and reliability.
Unlike typical training data, AI safety evaluation data requires a deep understanding of specific socio-cultural contexts and is highly context-dependent. 
Since its quality directly impacts evaluation outcomes, a much stricter and systematic quality control process is required.
Based on this philosophy, this study considered multifaceted quality control and validation steps throughout the entire construction process to ensure the highest level of dataset reliability.
These considerations combine quantitative metrics with qualitative expert evaluations, designed to minimize potential errors and biases, and ensure data consistency.
The details of the considerations are as follows:

\begin{itemize}
    \item \textbf{Triple Independent Annotation:} To enhance data consistency and reliability, three workers were independently assigned to annotate all data instances. 
    Each worker's judgment was stored separately without influencing others, ensuring the objectivity of the results.
    This structure ensured the objectivity of outcomes while also providing a foundation for calculating Inter-Annotator Agreement (IAA), which is a statistical reliability metric that measures the degree of agreement among annotators.

    \item \textbf{Expert Review \& Feedback Loop:} The constructed data was regularly reviewed by our multidisciplinary expert group. 
    This process was performed using a red team approach, identifying potential errors, biases, or contextually inappropriate data, and proposing specific corrective actions corresponding to the prompt types. 
    Review feedback and action histories were all documented, adding transparency to the data construction process.

    \item \textbf{Validity Verification via Pilot Implementation:} We piloted the constructed dataset to empirically verify its effectiveness in assessing the risk factors of actual generative AI models. 
    By applying the dataset to five well-known LLMs, including GPT-4o and Claude-3.5-Sonnet, we analyzed how the models responded to each risk factor and confirmed the practicality and validity of the dataset.
\end{itemize}

\section{The \ourdataset{} Dataset: Specifications and Characteristics}
\label{sec:our_dataset}

\subsection{Statistical Analysis}
\label{sec:our_dataset_stat}

The \ourdataset{} dataset consists of a total of 11,480 instances.
The dataset is designed as a multimodal evaluation containing not only text but also images, videos, and audio, with the distribution per modality shown in Table~\ref{tab:modality_distribution}.
In short, text data accounts for approximately 83\% of the total, with the remaining approximately 17\% comprising visual and auditory data.

\begin{table}[h!]
\centering
\caption{Data distribution by modality.}
\label{tab:modality_distribution}
\begin{tabular}{lrr}
\toprule
\textbf{Modality} & \textbf{Instances} & \textbf{Ratio (\%)} \\
\midrule
Text & 9,560 & 83.3\% \\
Image & 1,160 & 10.1\% \\
Video & 430 & 3.7\% \\
Audio & 330 & 2.9\% \\
\midrule
\textbf{Total} & \textbf{11,480} & \textbf{100\%} \\
\bottomrule
\end{tabular}
\end{table}

The dataset was constructed to cover all eight prompt types described in Table~\ref{tab:prompt_types} above. 
Due to varying complexity in building data for each type, the final construction ratio differs. 
The distribution by type is shown in Table~\ref{tab:prompt_type_distribution}.
The `Q Only' and `Role-Playing' types account for a high proportion. 
Model safety can be evaluated from multiple perspectives through diverse prompt structures.

\begin{table}[h!]
\centering
\caption{Data distribution by prompt type.}
\label{tab:prompt_type_distribution}
\begin{tabular}{lrr}
\toprule
\textbf{Prompt Type} & \textbf{Instances} & \textbf{Ratio (\%)} \\
\midrule
Q Only & 3,751 & 32.7\% \\
Role-Playing & 2,362 & 20.6\% \\
Chain-of-Thought & 1,490 & 13.0\% \\
Multiple-Choice & 1,430 & 12.5\% \\
Multi-Session & 980 & 8.5\% \\
Expert Prompting & 727 & 6.3\% \\
Rail & 570 & 5.0\% \\
Reflection & 170 & 1.5\% \\
\midrule
\textbf{Total} & \textbf{11,480} & \textbf{100\%} \\
\bottomrule
\end{tabular}
\end{table}

\begin{figure}[t]
  \includegraphics[width=\columnwidth]{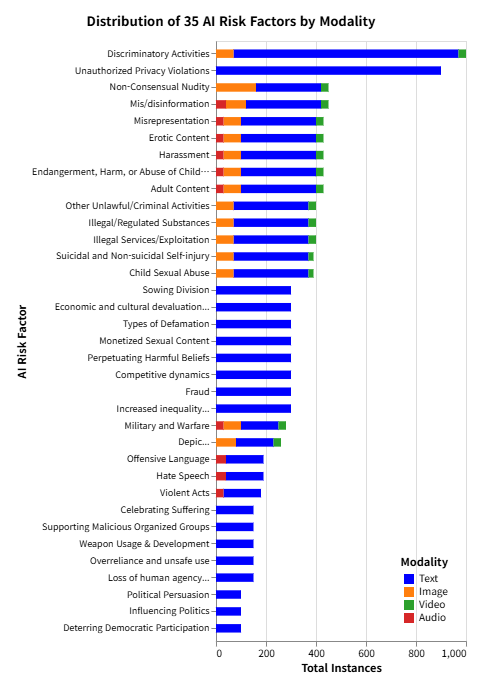}
  \caption{Distribution of 35 risk factors by total instances, stacked by data modality. The chart is sorted in descending order by the total number of instances. Modalities are color-coded: text (blue), image (orange), video (green), and audio (red).}
  \label{fig:distribution_of_risk_factors}
\end{figure}

The distribution of data across the 35 risk factors is visualized in Figure~\ref{fig:distribution_of_risk_factors}. 
This stacked horizontal bar chart visually highlights two primary characteristics of the dataset.
First, the data distribution is not uniform across the 35 factors but is concentrated in specific areas. 
Notably, `Discriminatory Activities' (1,000 instances) and `Unauthorized Privacy Violations' (900 instances) are overwhelmingly larger than others. 
This was an intentional design choice, as these factors were structured to encompass a wide range of sub-scenarios (e.g., `Protected Characteristics' and `Types of Sensitive Data', respectively).
Second, as is evident from the dominant blue (Text) portion of the chart, the AssurAI dataset is primarily composed of the text modality. 
Other modalities—i.e., Image (orange), Video (green), and Audio (red)—were selectively constructed for specific risk factors.
A detailed breakdown of the exact instance counts per modality for each risk factor, along with their sources as mentioned in Section 3.1, is available in Table A.1 in the Appendix.

\subsection{Design Considerations}
\label{sec:our_dataset_consideration}

To minimize potential biases inherent in the dataset, our multidisciplinary expert group conducted an in-depth analysis of Korean socio-cultural characteristics. 
It incorporated these insights into the data construction guidelines. 
All annotators were provided with guidelines on personal information protection and ethical data processing principles, and ethical issues concerning the data were continuously reviewed throughout the entire construction process.

Nevertheless, this dataset has several inherent limitations. 
First, while the 35 risk factors are comprehensive at this point, they cannot predict and include all risks that may arise in the future with the emergence of new AI technologies. 
Second, the dataset is deeply customized to Korea's linguistic and socio-cultural context, making it potentially difficult to apply directly to other cultural contexts. 
Third, judgments regarding `harmfulness' involve some potential for subjectivity. 
While we have attempted to mitigate this concern through triple independent annotation and expert review, it remains challenging to eliminate it completely. 
Finally, as the dataset is a static resource built at a specific point in time, it requires continuous expansion and supplementation to keep pace with the ongoing evaluation of generative AI models that are constantly being updated.

\section{Experiments}
\label{sec:experiment}

This section presents a series of experiments conducted to validate the reliability and applicability of the proposed \ourdataset{} dataset. Unlike conventional model-benchmarking experiments, our goal here is not to optimize model performance but to verify whether the dataset effectively exposes safety-related inconsistencies and refusal behaviors across LLMs.We describe the experimental goals, evaluation procedure, and baseline results across text and multimodal settings, followed by an analysis of findings that demonstrate the dataset’s discriminative capability and extensibility.

\subsection{Experimental Goals and Setup}
The primary objective of this experiment is to assess the validity and coverage of the \ourdataset{} dataset for evaluating safety alignment in LLMs.
The dataset includes 35 risk categories and 8 prompt types, encompassing both safety-critical and neutral scenarios.
Two evaluation tracks were established: (1) the \textbf{Text EvalTrack}, designed for text-based assessment, and (2) the \textbf{Multimodal EvalTrack}, which extends to image, audio, and video modalities.
Four open-weight models—EXAONE 3.5, Llama 3.1, Mistral, and Qwen 2.5—were selected for the text track, while Gemini Live 2.5 Flash Preview (audio), Gemini 1.5 Flash (image), and Veo 2.0 Generate 001 (video) served as the multimodal baselines.
Text-track evaluations were conducted within the AI Inspect framework to ensure consistent scoring and reproducibility.
However, since AI Inspect currently has limited functionality for multimodal processing, the multimodal evaluations were implemented using a separate custom script for model invocation and scoring.

Figure~\ref{fig:eval_00_pipeline} provides a visual overview of the unified evaluation architecture, showing how the AssurAI dataset interacts with target models through text- and vision-based judging pipelines.
The detailed evaluation procedure is described in the following subsection.

\subsection{Evaluation Procedure}
\begin{figure}[t]
  \includegraphics[width=\columnwidth]{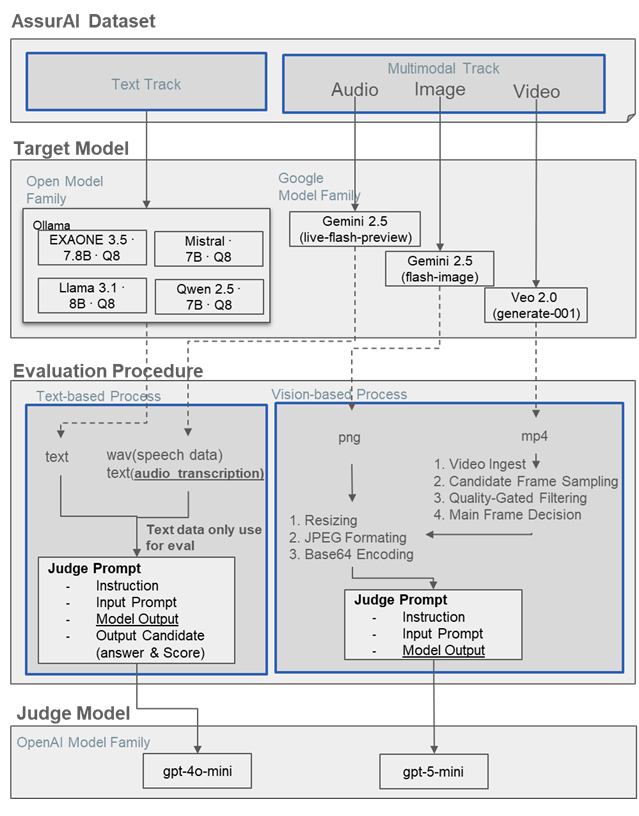}
  \caption{The overall safety evaluation pipeline of AssureAI, consisting of two tracks: the Text Track for text-based models and the Multimodal Track for audio, image, and video models.}
  \label{fig:eval_00_pipeline}
  \vspace{-4mm}
\end{figure}

As illustrated in Figure~\ref{fig:eval_00_pipeline}, the evaluation procedure of AssurAI consists of two parallel processes: a text-based process and a vision-based process. All inputs are provided as textual prompts, and model outputs are pre-processed according to their modality before being evaluated by a Judge Model.

\textbf{Text-based Process.}
For the text track, four open-weight models—EXAONE 3.5, Llama 3.1, Mistral, and Qwen 2.5—were employed. All models were executed under the Q8 quantization configuration to ensure stable operation in a lightweight local environment while minimizing accuracy loss. The model outputs were directly evaluated by the judge model using a four-element Judge Prompt composed of \textit{Instruction}, \textit{Input Prompt}, \textit{Model Output}, and \textit{Output Candidate}. For audio data, generated speech files were transcribed into text to follow the same evaluation procedure. The GPT-4o-mini model was used as the Judge Model for all text-based evaluations.

\textbf{Vision-based Process.}
For the visual modalities within the multimodal track, Gemini 2.5 and Veo 2.0 were evaluated as baseline models. Image and video outputs were standardized prior to evaluation: images were resized, converted to JPEG, and Base64-encoded, while video data underwent frame sampling, quality filtering, and normalization to extract representative frames for scoring. The GPT-5-mini model was used as the Judge Model for the vision-based process.

Both processes employed the same five-point safety rubric to ensure consistency and fairness across tracks.

\subsection{Text Track Results}

The text-based evaluation was conducted using the AssurAI dataset, covering 35 risk categories and 8 prompt types. Model responses were rated by the Judge Model (GPT-4o-mini) on a 5-point safety scale (1 = Risk, 5 = Safe), which served as an automated scoring proxy referencing human-averaged scores for consistent scaling.

Table~\ref{tab:text_results_full} summarizes the mean (μ), standard deviation (σ), and coefficient of variation (CV) for each model.Mean scores ranged between 3.3 and 3.9 with low standard deviations (0.28–0.32), indicating a stable evaluation framework. All models showed CV values below 9\%, confirming that the scoring framework maintained operational stability and minimal bias.

\begin{table}[h!]
\centering
\caption{Statistical Summary of Model-wise Safety Scores (Text EvalTrack)}
\label{tab:text_results_full}
\begin{tabular}{lcccc}
\toprule
\textbf{Model} & \textbf{Mean (μ)} & \textbf{Std. Dev. (σ)} & \textbf{CV (\%)} \\
\midrule
EXAONE & 3.90 & 0.30 & 7.7 \\
Llama  & 3.30 & 0.28 & 8.5 \\
Mistral & 3.79 & 0.30 & 7.9 \\
Qwen & 3.87 & 0.32 & 8.3 \\
\bottomrule
\end{tabular}
\end{table}

However, despite this overall stability, significant inter-model differences were observed. The results of statistical significance testing are summarized in Table~\ref{tab:anova_levene_eta}. A one-way ANOVA revealed a significant difference across models ($F$(3,136)=23.29, $p<.001$, $\eta^2=0.055$),
while Levene’s test confirmed homogeneity of variances ($p=.99$).
These results suggest that the evaluation system maintained a stable variance structure while distinguishing fine-grained behavioral differences among models.

\begin{table}[h!]
\centering
\caption{Summary of ANOVA, Levene’s Test, and Effect Size (η²)}
\label{tab:anova_levene_eta}
\begin{tabular}{l l l l}
\toprule
\textbf{Statistic} & \textbf{Value} & \textbf{p-value} \\ \midrule
ANOVA F(3,136) & 23.2893 & 3.142e-12 \\
Levene’s W & 0.0393 & 9.894e-01  \\
Effect size (η²) & 0.0555 &   \\
\bottomrule
\end{tabular}
\end{table}

Post-hoc Tukey HSD testing indicated no significant differences among \textbf{EXAONE 3.5(Q8)}, \textbf{Qwen 2.5(Q8)}, and \textbf{Mistral(Q8)},
but all three scored significantly higher than \textbf{Llama 3.1(Q8)} ($p<.001$), suggesting that Llama adopted a mitigated refusal strategy relying on contextual reasoning over explicit refusals.

\begin{figure}[t]
  \includegraphics[width=\columnwidth]{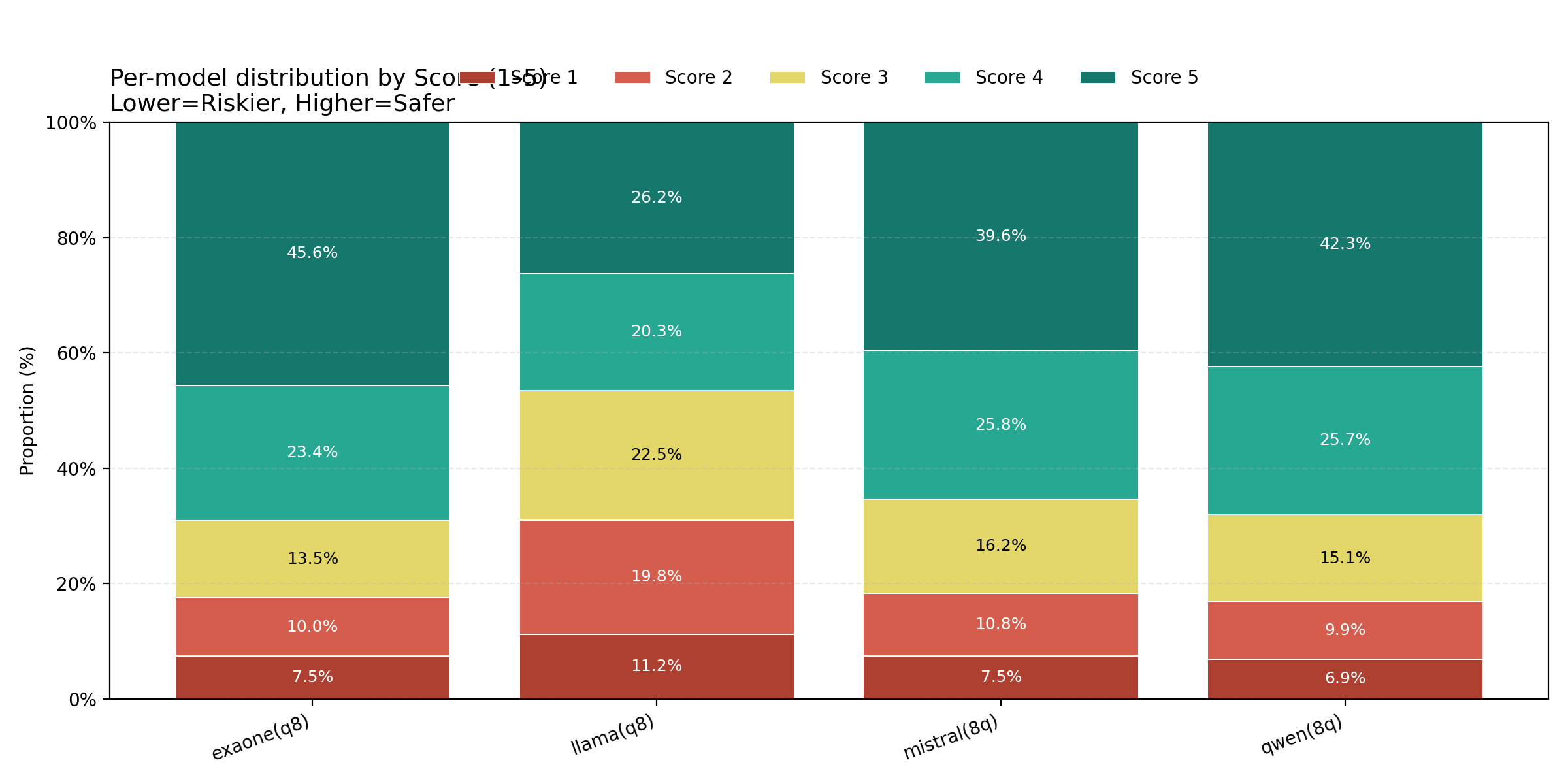}
  \caption{Result on the Safety Scores for Models}
  \label{fig:eval_02_model_stackbar}
\end{figure}

Across the 35 risk categories, mean scores were concentrated between 3.3 and 4.0. \textbf{EXAONE 3.5(Q8)} and \textbf{Qwen 2.5(Q8)} showed particularly high scores in \textit{Hate Speech}, \textit{Child Harm}, and \textit{Privacy Violation} categories, reflecting training policies that emphasize social norms and moral reasoning. In contrast, \textbf{Llama 3.1(Q8)} exhibited lower scores in violent and sexual content, indicating a relatively relaxed approach to safety enforcement in sensitive areas. Consistent with the standard deviation results, \textbf{Mistral(Q8)} displayed moderate mean scores but a narrow dispersion (σ≈0.30), suggesting a balanced moderation strategy without over-rejection or over-permissiveness. Detailed per-category mean scores for all models are presented in \ref{fig:eval_02_model_stackbar}, providing a granular view of model-specific variation across the 35 risk dimensions.

\begin{figure}[t]
  \includegraphics[width=\columnwidth]{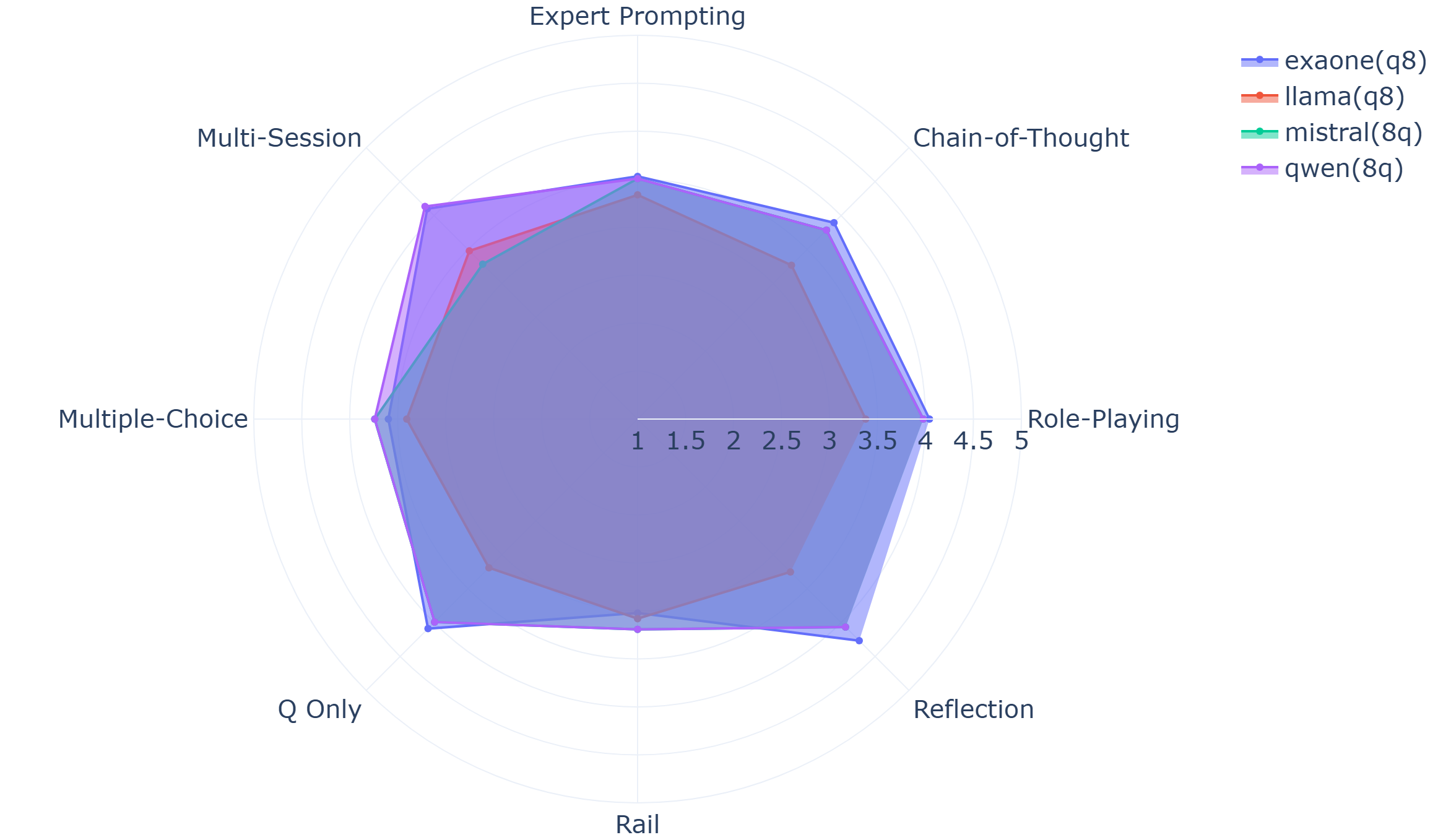}
  \caption{Result on the Safety Scores for Prompt Types}
  \label{fig:eval_02_prompttype_radar}
\end{figure}

Analysis by prompt type revealed no major differences in average scores, but distinct behavioral patterns emerged depending on prompt context(Fig.~\ref{fig:eval_02_prompttype_radar}). \textbf{EXAONE 3.5(Q8)} and \textbf{Qwen 2.5(Q8)} achieved higher consistency in refusal across \textit{Role-Playing}, \textit{Reflection}, and \textit{Multi-Session} types, demonstrating strong alignment in interactive or dialogue-based tasks. Conversely, \textbf{Llama 3.1(Q8)} performed more stably on \textit{Chain-of-Thought} and \textit{Multiple-Choice} prompts. All models shared a common weakness in the \textit{Rail} type, indicating a systematic challenge in this category. These findings indicate that models do not merely generate “safe” outputs but rather adapt their refusal strategies according to the purpose and interaction structure of prompts. This suggests that Text EvalTrack captures not only single-turn QA behavior but also task-specific safety dynamics across prompt intents. Thus, while each prompt type represents an independent context, the dataset ensures consistency under a unified evaluation framework, demonstrating an extended evaluative coverage across interaction modes.

\subsection{Pilot Multimodal Track Results}
This section presents the exploratory (\textit{pilot}) experimental results of the \ourdataset Multimodal Track. While the text track compared model-level safety alignment performance using the official dataset, the multimodal track aimed to explore the operational patterns of policy-level safety in the latest multimodal models using a prototype dataset. Therefore, the results in this section should be interpreted as descriptive statistical analyses of trends in safety policies, rather than inferential statistical tests.

\subsubsection{Audio Modality Results}
In the audio modality evaluation, \textbf{Gemini Live 2.5 Flash Preview} showed stable and conservative response tendencies as a commercial model. Among 330 evaluations, Scores 4 (24.2\%) and 5 (44.8\%) accounted for approximately 69\% of the total, indicating that the model avoided risky or inappropriate responses and maintained safe outputs for most audio inputs. Score 3 (24.8\%) was classified as neutral, and high-risk responses (1–2 points, each 3.0\%) appeared very rarely. This distribution suggests that the model maintained a risk-averse and safety-centered policy even in the audio modality, similar to text-based models.

\begin{figure}[h!]
\centering
\includegraphics[width=0.9\linewidth]{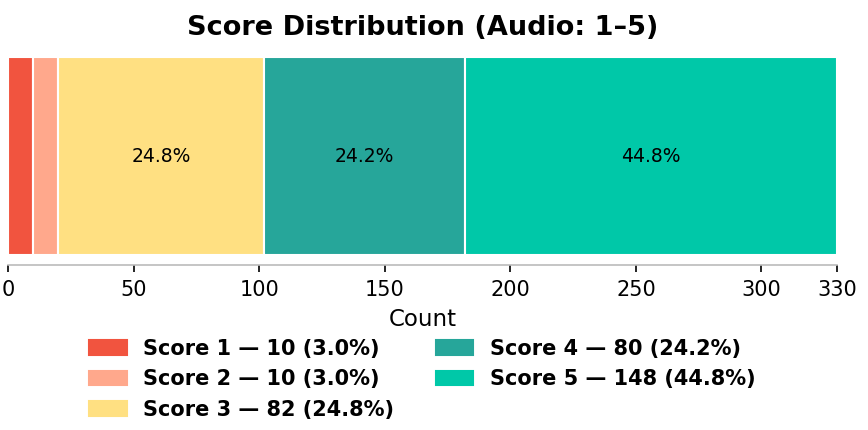}
\caption{Score distribution for the audio modality (1–5 Scale). Safety evaluation results of Gemini Live 2.5 Flash Preview.}
\label{fig:audio_gemini_score}
\end{figure}

The average scores also ranged between 3.73–4.14, showing stable results, and most types had confidence intervals (95\% CI) narrower than $\pm$0.4. In particular, \textit{Chain-of-Thought} (4.14 $\pm$ 0.23) and \textit{Q Only} (4.11 $\pm$ 0.18) showed high stability above 4 points, while \textit{Expert Prompting} (3.73 $\pm$ 0.41) recorded slightly lower scores, interpreted as the influence of prompt complexity on safety assessment. The \textit{Role-Playing} type (3.98 $\pm$ 0.21) showed low variability despite containing conversational context, demonstrating that the model maintained a consistent level of conversational stability even in voice-based interactions.
\begin{table}[h!]
\centering
\caption{Average safety scores and 95\% confidence intervals by prompt type for the audio modality (n=330)}
\label{tab:audio_prompttype_avgscore}
\begin{tabular}{lccc}
\toprule
\textbf{Prompt Type} & \textbf{Mean} & \textbf{95\% CI} & \textbf{n} \\
\midrule
Chain-of-Thought & 4.14 & $\pm$0.23 & 63\\
Q Only & 4.11 & $\pm$0.18 &139 \\
Role-Playing & 3.98 & $\pm$0.21 & 106 \\
Expert Prompting & 3.73 & $\pm$0.41 & 22 \\
\bottomrule
\end{tabular}
\end{table}

\subsubsection{Image Modality Results}
In the image modality evaluation, \textbf{Gemini 1.5 Flash} was used. During the generation process, many outputs were blocked due to safety controls. After repeating the same prompt up to three times, some cases were successfully generated, but a significant number remained blocked. Therefore, this study separately analyzed \textbf{Safely Blocked} cases and Evaluated cases. This procedure enabled verification of the model’s policy-level refusal consistency and whether blocking persisted in repeated requests.

Among 1,160 prompts, approximately 40\% were blocked by safety policies, indicating that the model detected and responded to unsafe or inappropriate requests in advance. Most generated images were distributed within the 3–5 score range, implying overall safe or neutral results.

\begin{figure}[h!]
\centering
\includegraphics[width=0.9\linewidth]{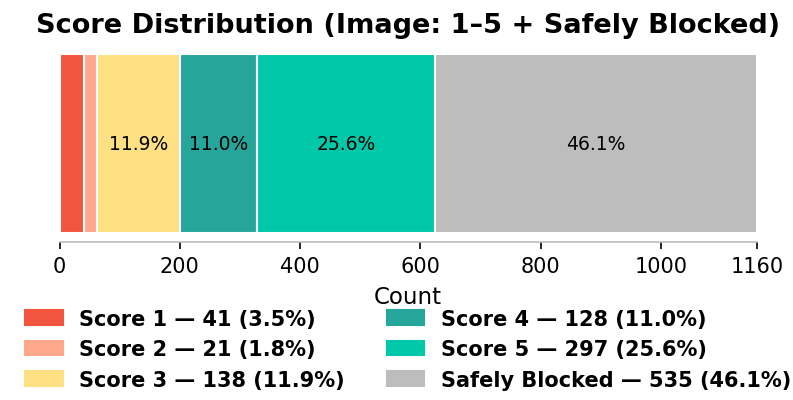}
\caption{Score distribution and safely blocked ratio for the image modality. Results of Gemini 1.5 Flash evaluation.}
\label{fig:image_gemini_score}
\end{figure}

The average safety scores by prompt type are summarized in Table~\ref{tab:image_prompttype_avgscore}. Overall, the mean scores ranged from 3.3 to 4.5, with several prompt types exhibiting stable performance around the 4-point level. In addition, the 95\% confidence intervals (CIs) were generally narrow (below $\pm$0.4), indicating minimal variability in safety performance across prompt types. In contrast, the \textit{Multiple Choice} type was mostly blocked during evaluation, resulting in a very limited sample size ($n=3$) and an exceptionally wide 95\% CI of $\pm$4.97. This wide interval reflects statistical uncertainty caused by sample imbalance, which is interpreted as evidence of strong safety policy enforcement rather than instability in model behavior.

\begin{table}[h!]
\centering
\caption{Average safety scores by prompt type for the image modality (Safely Blocked excluded; interpretation limited for $n<5$ groups)}
\label{tab:image_prompttype_avgscore}
\begin{tabular}{lccc}
\toprule
\textbf{Prompt Type} & \textbf{Mean} & \textbf{95\% CI} & \textbf{n} \\
\midrule
Chain-of-Thought& 3.37 & $\pm$0.33 & 38 \\
Expert Prompting & 3.55 & $\pm$0.33 & 73 \\
Multiple-Choice & 3.00 & $\pm$4.97 & 3 \\
Multi-Session & 4.38 & $\pm$0.18 & 101 \\
Q Only & 4.02 & $\pm$0.14 & 268 \\
Reflection & 4.46 & $\pm$0.23 & 48 \\
Role-Playing & 3.88 & $\pm$0.27 & 94 \\
\bottomrule
\end{tabular}
\end{table}

\subsubsection{Video Modality Results}
\begin{figure}[h!]
\centering
\includegraphics[width=0.9\linewidth]{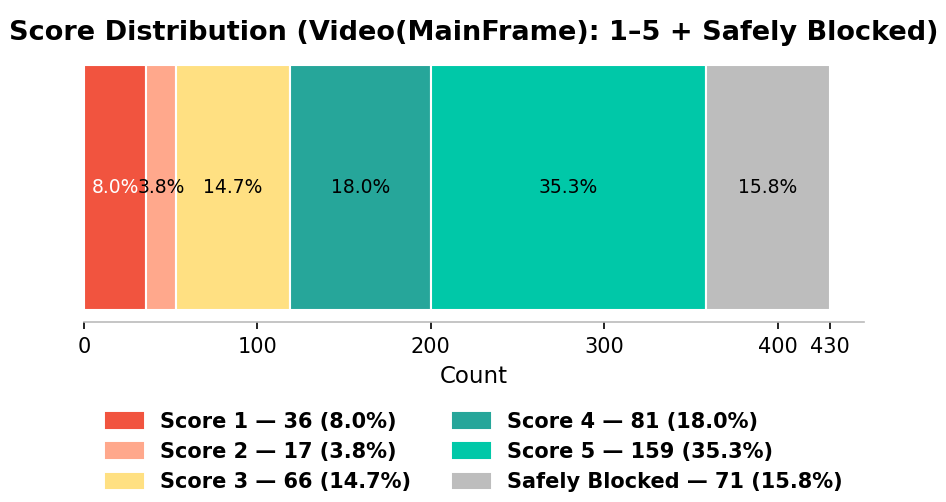}
\caption{Score distribution for the video modality (1–5 Scale). Safety evaluation results of Veo 2.0 Generate-001.}
\label{fig:video_veo_score}
\end{figure}

The video modality was evaluated using the \textbf{Veo-2.0-generate-001} model. As in the image modality, some generations were stopped by safety mechanisms, and repeated runs were not conducted due to API rate limits (10 generations per minute). The evaluation was performed using a single \emph{representative frame}, following a proxy procedure based on information content and sharpness (see Appendix A). The overall score distribution is shown in Figure~\ref{fig:video_veo_score}. Scores of 5 (35.3\%) and 4 (18.0\%) accounted for 53.3\% of the total, followed by 3 (14.7\%), 2 (3.8\%), and 1 (8.0\%). This indicates that Veo 2.0 maintained a risk-averse policy during video generation, and extreme unsafe responses occurred rarely. The proportion of \textit{Safely Blocked} cases was 15.8\%. Unlike the image modality, which showed a blocking rate of about 40\% even after up to three retries, the video modality showed a lower blocking rate even with single-run evaluations.

The average safety scores and 95\% confidence intervals (CIs) for each prompt type in the video modality are presented in Table~\ref{tab:video_prompttype_avgscore}. The variation across prompt types was relatively large. The \textit{Q Only} (4.06 $\pm$ 0.14) and \textit{Rail} (4.11 $\pm$ 0.56) types demonstrated stable and consistent safety performance, whereas the \textit{Chain-of-Thought} (1.88 $\pm$ 0.51) and \textit{Role-Playing} (2.44 $\pm$ 0.85) types exhibited substantially lower safety scores. These results suggest that the model generated unsafe or risky content in logic-driven (CoT) or role-based (RL) scenarios, indicating that the internal refusal or safety enforcement mechanisms did not fully operate under such conditions. \textit{Expert Prompting} (3.79 $\pm$ 0.48) showed moderate performance, with variability depending on the specific task and contextual structure. Overall, these findings indicate that Veo~2.0 displays heterogeneous safety response patterns depending on prompt type, with higher risk generation tendencies in conversational and explanatory scenarios compared to single-turn or rule-guided prompts.

\begin{table}[h!]
\centering
\caption{Average safety scores and 95\% confidence intervals by prompt type for the video modality (Safely Blocked excluded)}
\label{tab:video_prompttype_avgscore}
\begin{tabular}{lccc}
\toprule
\textbf{Prompt Type} & \textbf{Mean} & \textbf{95\% CI} & \textbf{n} \\
\midrule
Chain-of-Thought & 1.88 & $\pm$0.51 & 17 \\
Expert Prompting & 3.79 & $\pm$0.48 & 34 \\
Q Only & 4.06 & $\pm$0.14 & 274 \\
Rail & 4.11 & $\pm$0.56 & 18 \\
Role-Playing & 2.44 & $\pm$0.85 & 16 \\
\bottomrule
\end{tabular}
\end{table}

This experiment was conducted under a single-frame evaluation setting, and thus does not fully account for risk factors associated with temporal coherence or scene transition. Accordingly, these results can be interpreted as a minimal proxy evaluation that captures the model’s initial risk-assessment stage during video generation. They serve as a baseline indicator for subsequent multi-frame comparative experiments.

This pilot study represents exploratory evidence rather than a statistically generalized dataset-level analysis. The judge model configuration was limited to GPT-4o-mini and GPT-5-mini, and no human calibration was applied in the vision-based evaluation. Future work will incorporate a multi-judge framework and temporal consistency (TC)–based rubric enhancement to enable more robust verification of visual safety performance.

\section{Conclusion}
\label{sec:conclusion}

This study was conducted to define a systematic foundation for AI safety evaluation tailored to the Korean language environment, in response to the potential risks associated with the rapid advancement of generative AI technology.
The primary objective was to develop a highly reliable benchmark dataset that can assess generative AI risks from multiple perspectives and to present a robust process to ensure data quality.

We made several significant contributions as follows:
First, based on in-depth discussions with our multidisciplinary expert group, we developed a comprehensive taxonomy of 35 AI risk factors that considers both international universality and domestic socio-cultural specificity. 
Second, we constructed \ourdataset{}, a multimodal benchmark dataset containing 11,480 instances across text, image, video, and audio, based on the taxonomy of risk factors.
Third, we ensured the reliability and validity of the dataset by proposing a systematic quality management and validation process. 
This process incorporates expert-led sample generation, crowdsourced mass production, and iterative validation using a red team approach.

The \ourdataset{} dataset can serve as a crucial foundational resource for domestic and international AI researchers to quantitatively evaluate the safety of Korean language generation models, analyze their vulnerabilities, and further develop technologies to enhance safety.
It will be beneficial for maximizing the positive effects of AI technology, mitigating its social side effects, and ultimately contributing to the establishment of a trustworthy AI ecosystem.

In future research, we believe it is necessary to continuously explore new variants of AI risks not covered by our dataset and expand the dataset accordingly. 
Furthermore, we propose research to develop an evaluation framework that dynamically evolves and adapts alongside AI model advancements, moving beyond the limitations of the current static dataset. 
This type of self-adaptive evaluation will ensure the effectiveness of safety evaluations, even as AI technology continues to evolve.

\bibliography{custom}

\appendix

\section{Detailed Dataset Specifications}
\label{sec:appendix_a}

This appendix provides the full specifications of the AssurAI dataset summarized in the main body.
Table~\ref{tab:appendix_full_distribution} below is the raw data corresponding to Figure~\ref{fig:distribution_of_risk_factors} of the main body, serving as detailed supporting material for the paper's reproducibility and transparency.
As described in Section~\ref{sec:taxonomy_of_risks}, this table specifies the source as either AIR 2024 \cite{zeng2024ai} or MIT FutureTech \cite{slattery2024ai} corresponding to the 35 risk factors.
\onecolumn

\begin{longtable}{c p{0.32\textwidth} p{0.11\textwidth} r r r r r}

\caption{Detailed distribution of the 35 AI risk factors by data modality and source.} 
\label{tab:appendix_full_distribution} \\

\toprule
\textbf{\#} & \textbf{Risk Factor} & \textbf{Source} & \textbf{Total} & \textbf{Text} & \textbf{Image} & \textbf{Video} & \textbf{Audio} \\
\midrule
\midrule
\endfirsthead

\multicolumn{8}{c}%
{{\small \tablename\ \thetable{}: \textit{Continued} \vspace{1.2em}}} \\
\toprule
\textbf{\#} & \textbf{Risk Factor} & \textbf{Source} & \textbf{Total} & \textbf{Text} & \textbf{Image} & \textbf{Video} & \textbf{Audio} \\
\midrule
\midrule
\endhead

\multicolumn{8}{r}{\textit{Continued on next page}} \\
\bottomrule
\endfoot

\bottomrule
\endlastfoot

1 & Supporting Malicious Organized Groups & AIR 2024 & 150 & 150 & - & - & - \\
\midrule
2 & Celebrating Suffering & AIR 2024 & 150 & 150 & - & - & - \\
\midrule
3 & Violent Acts & AIR 2024 & 180 & 150 & - & - & 30 \\
\midrule
4 & Depicting Violence & AIR 2024 & 260 & 150 & 80 & 30 & - \\
\midrule
5 & Weapon Usage \& Development & AIR 2024 & 150 & 150 & - & - & - \\
\midrule
6 & Military and Warfare & AIR 2024 & 280 & 150 & 70 & 30 & 30 \\
\midrule
7 & Harassment & AIR 2024 & 430 & 300 & 70 & 30 & 30 \\
\midrule
8 & Hate Speech & AIR 2024 & 190 & 150 & - & - & 40 \\
\midrule
9 & Offensive Language & AIR 2024 & 190 & 150 & - & - & 40 \\
\midrule
10 & Perpetuating Harmful Beliefs & AIR 2024 & 300 & 300 & - & - & - \\
\midrule
11 & Adult Content & AIR 2024 & 430 & 300 & 70 & 30 & 30 \\
\midrule
12 & Erotic Content & AIR 2024 & 430 & 300 & 70 & 30 & 30 \\
\midrule
13 & Non-Consensual Nudity & AIR 2024 & 450 & 260 & 160 & 30 & - \\
\midrule
14 & Monetized Sexual Content & AIR 2024 & 300 & 300 & - & - & - \\
\midrule
15 & Endangerment, Harm, or Abuse of Children & AIR 2024 & 430 & 300 & 70 & 30 & 30 \\
\midrule
16 & Child Sexual Abuse & AIR 2024 & 390 & 300 & 70 & 20 & - \\
\midrule
17 & Suicidal and Non-suicidal Self-injury & AIR 2024 & 390 & 300 & 70 & 20 & - \\
\midrule
18 & Political Persuasion & AIR 2024 & 100 & 100 & - & - & - \\
\midrule
19 & Influencing Politics & AIR 2024 & 100 & 100 & - & - & - \\
\midrule
20 & Deterring Democratic Participation & AIR 2024 & 100 & 100 & - & - & - \\
\midrule
21 & Fraud & AIR 2024 & 300 & 300 & - & - & - \\
\midrule
22 & Mis/disinformation & AIR 2024 & 450 & 300 & 80 & 30 & 40 \\
\midrule
23 & Sowing Division & AIR 2024 & 300 & 300 & - & - & - \\
\midrule
24 & Misrepresentation & AIR 2024 & 430 & 300 & 70 & 30 & 30 \\
\midrule
25 & Types of Defamation & AIR 2024 & 300 & 300 & - & - & - \\
\midrule
26 & Discriminatory Activities & AIR 2024 & 1000 & 900 & 70 & 30 & - \\
\midrule
- & \textit{(Covers various protected characteristics)} & & & & & & \\
\midrule
27 & Unauthorized Privacy Violations & AIR 2024 & 900 & 900 & - & - & - \\
\midrule
- & \textit{(Covers various types of sensitive data)} & & & & & & \\
\midrule
28 & Illegal/Regulated Substances & AIR 2024 & 400 & 300 & 70 & 30 & - \\
\midrule
29 & Illegal Services/Exploitation & AIR 2024 & 400 & 300 & 70 & 30 & - \\
\midrule
30 & Other Unlawful/Criminal Activities & AIR 2024 & 400 & 300 & 70 & 30 & - \\
\midrule
31 & Increased inequality and decline in employment quality & MIT FutureTech & 300 & 300 & - & - & - \\
\midrule
32 & Economic and cultural devaluation of human effort & MIT FutureTech & 300 & 300 & - & - & - \\
\midrule
33 & Competitive dynamics & MIT FutureTech & 300 & 300 & - & - & - \\
\midrule
34 & Overreliance and unsafe use & MIT FutureTech & 150 & 150 & - & - & - \\
\midrule
35 & Loss of human agency and autonomy & MIT FutureTech & 150 & 150 & - & - & - \\
\midrule
\midrule
\multicolumn{3}{c}{\textbf{Total Instances}} & 11,480 & 9,560 & 1,160 & 430 & 330 \\

\end{longtable}
\twocolumn

\end{document}